\documentclass[letterpaper]{article} 
\usepackage{aaai2026} 
\usepackage{times}  
\usepackage{helvet}  
\usepackage{courier}  
\usepackage[hyphens]{url}  
\usepackage{graphicx} 
\usepackage{natbib}  
\usepackage{caption} 
\usepackage{algorithm}
\usepackage{algorithmic}
\usepackage{multirow}
\usepackage{booktabs}
\usepackage{array}
\usepackage{colortbl}
\usepackage{xcolor}
\usepackage{amsmath}
\usepackage{pifont}
\usepackage{amssymb}
\usepackage{amsthm}
\newtheorem{theorem}{Proposition}
\usepackage{newfloat}
\usepackage{listings}

\DeclareCaptionStyle{ruled}{labelfont=normalfont,labelsep=colon,strut=off} 
\floatstyle{ruled}
\newfloat{listing}{tb}{lst}{}
\floatname{listing}{Listing}
\title{Fine-Tuned LLMs Know They Don't Know: \\ A Parameter-Efficient Approach to Recovering Honesty}
\author{
    Zeyu Shi\textsuperscript{\rm 1}\thanks{\ \ Equal contribution.}, 
    Ziming Wang\textsuperscript{\rm 1}\footnotemark[1], 
    Tianyu Chen\textsuperscript{\rm 1}\thanks{\ \ Corresponding author.}, 
    Shiqi Gao\textsuperscript{\rm 1}, \\
    Haoyi Zhou\textsuperscript{\rm 2,3}\footnotemark[2], 
    Qingyun Sun\textsuperscript{\rm 1}, 
    Jianxin Li\textsuperscript{\rm 1,3} \\
}
\affiliations{

    \textsuperscript{\rm 1}SKLCCSE, School of Computer Science and Engineering, Beihang University\\
    \textsuperscript{\rm 2}School of Software, Beihang University\\
    \textsuperscript{\rm 3}Zhongguancun Laboratory, Beijing\\
    \{szy\_629, wangzm412, tianyuc, gaoshiqi, haoyi, lijx\}@buaa.edu.cn
}

\usepackage{bibentry}

\begin{document}

\maketitle

\begin{abstract}
The honesty of Large Language Models (LLMs) is increasingly important for safe deployment in high-stakes domains. However, this crucial trait is severely undermined by supervised fine-tuning (SFT), a common technique for model specialization. Existing recovery methods rely on data-intensive global parameter adjustments, implicitly assuming that SFT deeply corrupts the models' ability to recognize their knowledge boundaries. However, we observe that fine‑tuned LLMs still preserve this ability; what is damaged is their capacity to faithfully express that awareness. Building on this, we propose Honesty-Critical Neurons Restoration (HCNR) to surgically repair this suppressed capacity. HCNR identifies and restores key expression-governing neurons to their pre-trained state while harmonizing them with task-oriented neurons via Hessian-guided compensation. Experiments on four QA tasks and five LLM families demonstrate that HCNR effectively recovers 33.25\% of the compromised honesty while achieving at least 2.23x speedup with over 10x less data compared to baseline methods, offering a practical solution for trustworthy LLM deployment.
\end{abstract}


\section{Introduction}

As Large Language Models (LLMs) are increasingly integrated into high-stakes domains~\cite{zhang2024catastrophic,sarabadani2019detection}, their reliability is not merely a feature but a necessity~\cite{2021A}. A cornerstone of this reliability is honesty, which has two components~\cite{li2024survey}. The first is self-knowledge: the ability to recognize their knowledge boundaries and distinguish what they know from what they do not. The second is faithful self-expression based on this awareness. This trait is crucial because LLMs that confidently fabricate facts or recommend false cures can cause serious harm, undermining user trust and safety.

The honesty of LLMs is typically instilled during the alignment stage through techniques such as Reinforcement Learning from Human Feedback (RLHF)~\cite{glaese2022improving}, which enable models to refuse inappropriate questions or those beyond their knowledge boundaries~\cite{bai2022training,manish2023constitutional}. However, this acquired honesty is not immutable. Recent researches find that the supervised fine-tuning (SFT) could greatly hurt the honesty of LLMs, such as in legal QA~\cite{dahl2024large}, medical diagnosis~\cite{kim2025medical} and educational content generation~\cite{nguyen2025smoothing}. To recover the honesty of LLM after SFT, previous methods ~\cite{zhang2023r,li2024survey,cheng2024can} interfere heavily with global parameters using extensive datasets, under the assumption that the model's knowledge boundaries have been deeply corrupted and its capacity for self-knowledge has been lost. 

\begin{figure}[t]
    \centering
    \includegraphics[width=\linewidth]{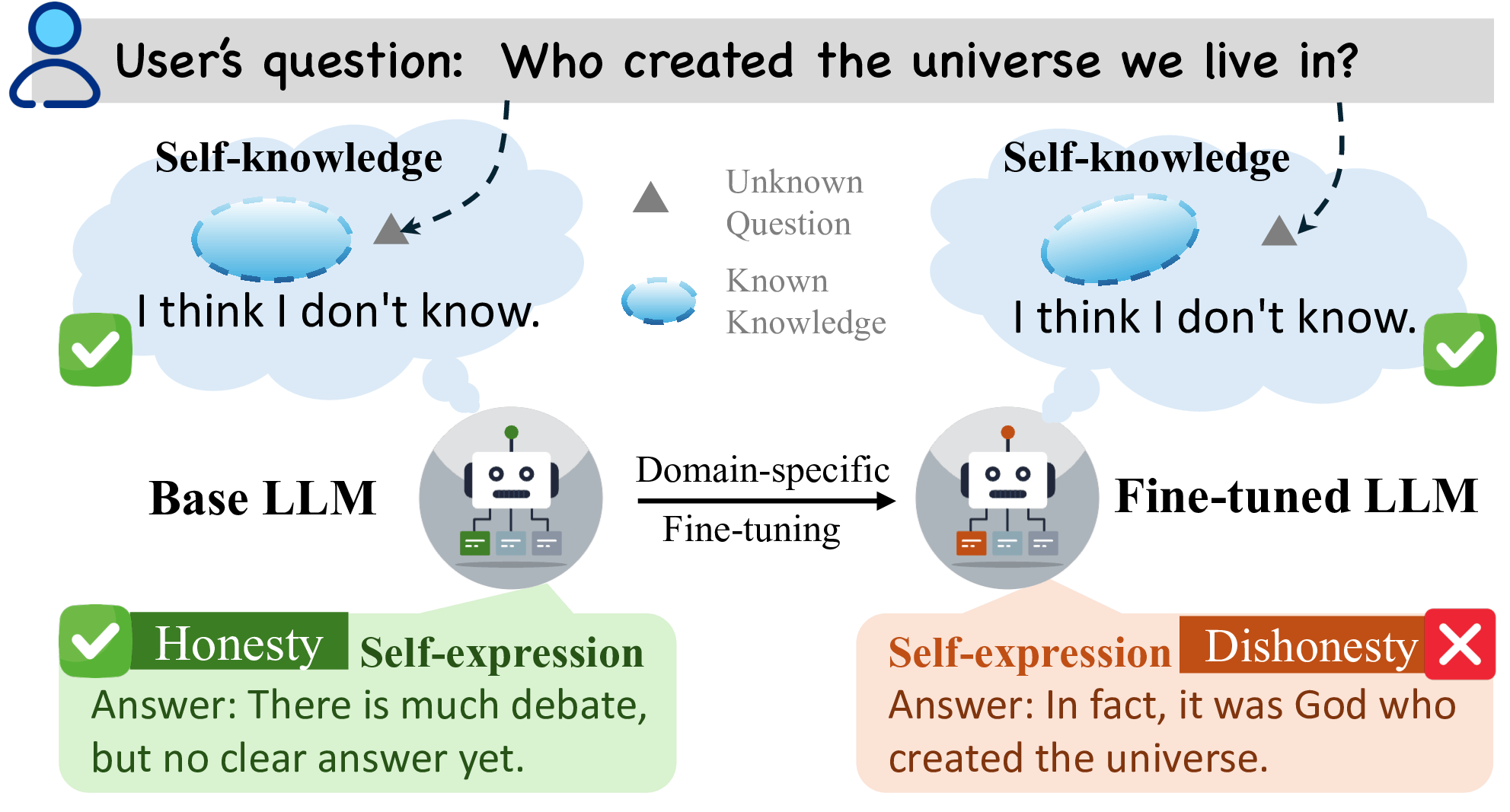}
    \caption{Mechanism of honesty degradation in domain-specific fine-tuning. The dishonest behavior of a fine-tuned LLM arises from impaired self-expression, rather than a loss of self-knowledge, which remains intact. This understanding motivates our methods for honesty recovery.}
    \label{fig:intro}
\end{figure}

However, we observe that the failure of self-expression instead of self-knowledge is the root cause of fine-tuned LLMs' dishonesty (as illustrated in Figure \ref{fig:intro}), thus global interventions may not be necessary.
In this paper, we propose a targeted, parameter-efficient solution to recover the honesty, named Honesty-Critical Neurons Restoration (HCNR). By restoring neurons governing honesty expression to their pre-trained values, and harmonizing them with task-oriented neurons via Hessian-guided compensation, HCNR recovers honesty with high parameter efficiency and negligible task performance degradation. Our main contributions can be summarized as follows:

\begin{itemize}
    \item As far as we know, we are the first to reveal that the dishonesty induced by SFT is a spurious phenomenon, stemming not from a loss of self-knowledge but from the failure of self-expression.

    \item We propose the HCNR framework, a parameter-efficient honesty recovery framework, recalibrating the critical neurons to recover fine-tuned LLMs' honesty without sacrificing task performance.

    \item Extensive experiments prove that HCNR attains honesty recovery performance comparable to baseline methods without compromising downstream task performance, demonstrating remarkable efficiency by achieving at least a 2.23x speedup with over 10x less data.
\end{itemize}
    
\begin{figure}[b]
    \centering
    \includegraphics[width=\linewidth]{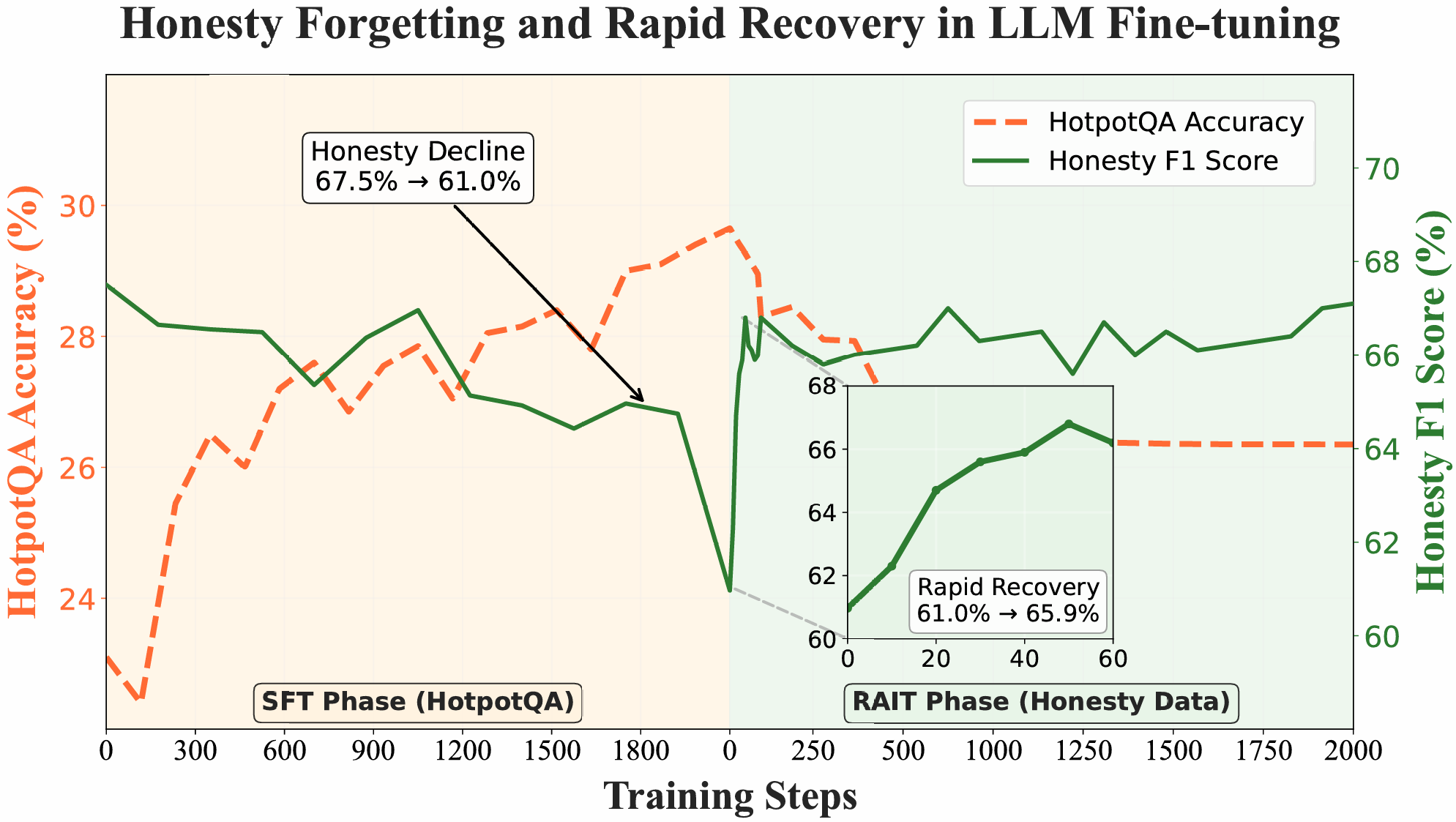}
        \caption{Trends in downstream performance and honesty during Domain SFT and RAIT: honesty declines substantially during Domain SFT, whereas under RAIT it rebounds sharply after only 60 gradient steps.}
    \label{fig:llm_honesty_recovery}
\end{figure}
\section{Understanding Honesty Degradation: LLMs’ Spurious Dishonesty}
Supervised fine-tuning (SFT) often degrades the honesty of large language models (LLMs), causing them to fabricate plausible yet fallacious responses to questions outside their knowledge scope, rather than expressing uncertainty. This behavior prompts a fundamental inquiry: \textit{Does this decline in honesty arise from a corruption of the model's inner awareness of its knowledge boundaries, or merely from an inability to articulate this awareness?}
\begin{figure}[h]
    \centering
    \includegraphics[width=\linewidth]{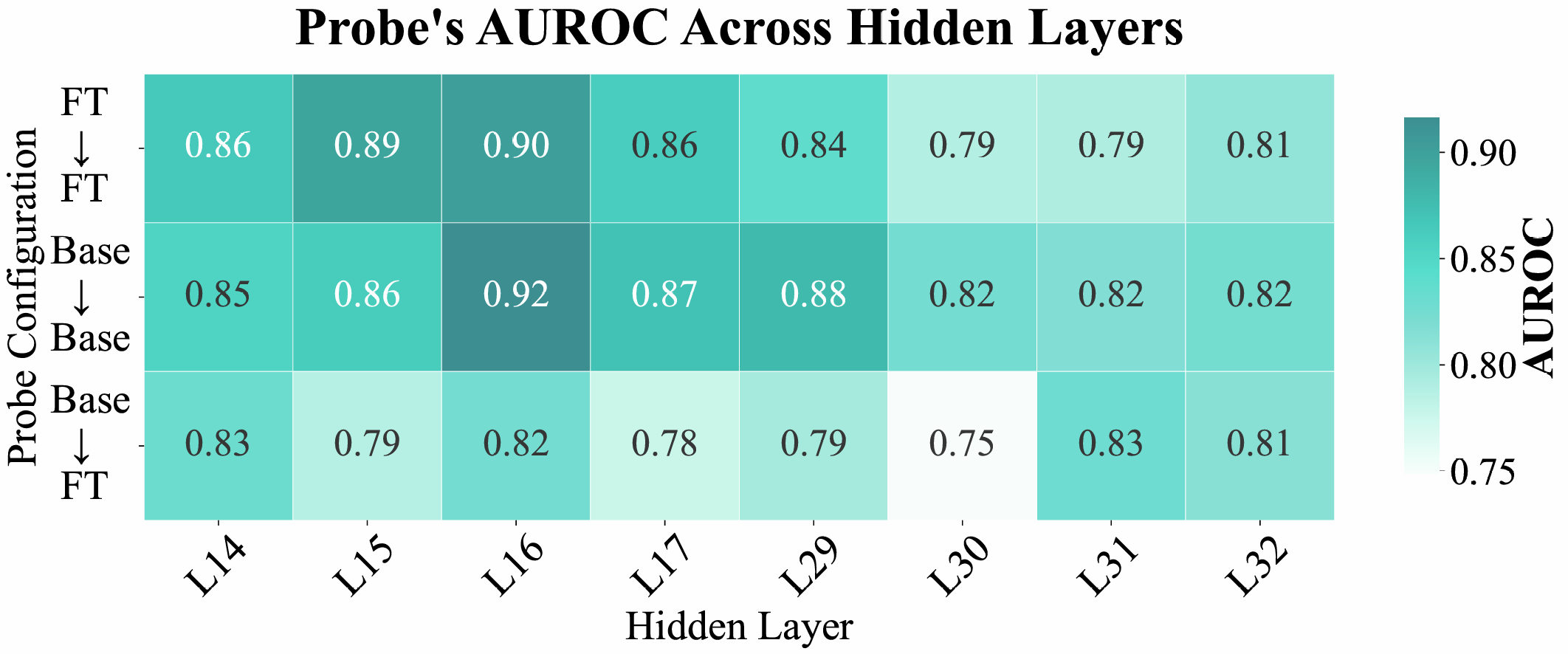}
    \caption{Logistic Regression probe's AUROC for distinguishing answerable vs. unanswerable. For brevity, base LLM is ``Base", fine-tuned LLM is ``FT". Row 1: Probes trained on the fine‑tuned LLM achieve high AUROC, confirming that knowledge‑boundary signals remain linearly separable. Rows 2–3: Probes trained on the base LLM preserve high AUROC when applied to the fine‑tuned model, demonstrating that SFT‑induced parameter shifts do not alter the geometric structure of these representations.}
    \label{fig:llm_probe_heatmap}
\end{figure}

\noindent \textbf{Experimental Setup.} To investigate this question, we fine‑tune Llama‑3.1‑8B‑Instruct with Low‑Rank Adaptation~\cite{hu2022lora} and full fine-tuning on the HotpotQA, yielding a fine‑tuned model whose honesty is markedly diminished. We then assess honesty using the FalseQA benchmark and curate a specialized dataset $D^{\text{hon}}$ for honesty recovery via RAIT~\cite{zhang2023r} (see Appendix B).

\noindent \textbf{Observation 1: Honesty Snaps Back in Few Steps.} Our initial investigation involves conducting honesty‑augmented training on the fine-tuned LLM using $D^{\text{hon}}$. As demonstrated in Figure~\ref{fig:llm_honesty_recovery}, the model's honesty recovers substantially after merely 60 gradient updates. This rapid recuperation suggests that the model's core knowledge boundary capabilities may remain intact, with honesty degradation potentially stemming from disrupted expression rather than impaired awareness~\cite{zhang2024dissecting, mai2024fine}.

\begin{figure*}[t]
    \centering
    \includegraphics[width=\linewidth]{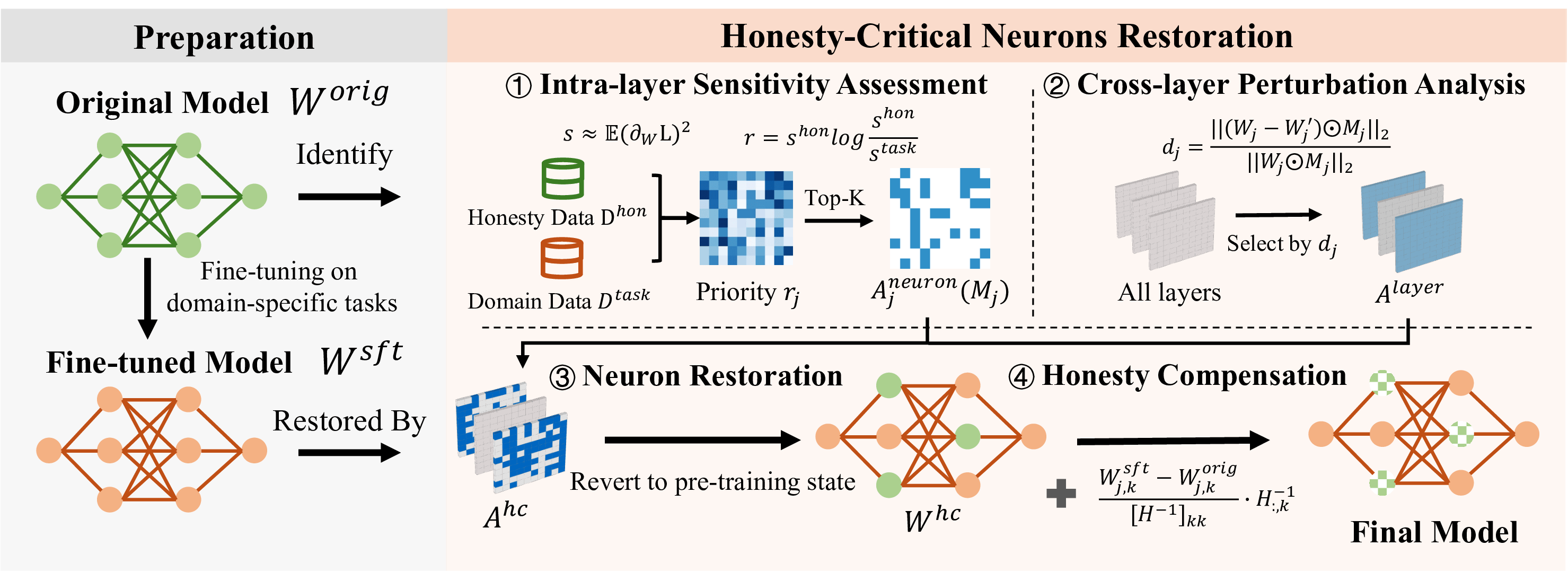}
    \caption{Honesty-Critical Neurons Restoration (HCNR) framework comprises two stages: In Stage 1, \ding{172} we first identify neurons whose Fisher-based importance is high for honesty but low for downstream tasks, \ding{173} then select from these candidates the neurons most severely perturbed by SFT, and \ding{174} subsequently restore these neurons to their pre-training states. In Stage 2, \ding{175} we employ a Hessian-guided compensation vector that makes minimal, targeted adjustments to these restored parameters, realigning them with task-oriented neurons and preventing collateral honesty loss.}
    \label{fig:overview}
\end{figure*}
\noindent \textbf{Observation 2: Boundary Signals Unshaken by SFT.} To substantiate this behavioral evidence with mechanistic insights, we examine whether the fine-tuned LLM's inner representations retain distinct signals for answerable versus unanswerable information. Specifically, we train linear probes (Logistic Regression classifiers) on hidden states extracted from layers of the fine-tuned LLM on FalseQA. Row 1 in Figure~\ref{fig:llm_probe_heatmap} reveals that these probes achieve high classification performance across all model layers, confirming that the fine-tuned LLM maintains a linearly separable representational structure that encodes knowledge boundaries.

To quantify the precise impact of SFT on these boundary-distinguishing signals, we employ a transfer learning paradigm. Specifically, we train linear probes on the base LLM's hidden states and apply them directly to the fine-tuned LLM's representations without retraining. The transferred probes maintain high AUROC scores (Rows 2-3 in Figure~\ref{fig:llm_probe_heatmap}), indicating the fundamental geometric structure that separates representations of known versus unknown information is robustly preserved during SFT. This phenomenon is likewise observed in LLMs fine‑tuned with full fine-tuning (details are shown in Appendix C).

These findings collectively demonstrate that \textbf{honesty degradation after SFT stems not from the destruction of knowledge boundary awareness, but rather from an impaired ability to express this preserved awareness}, as the underlying neural representations defining knowledge boundaries remain remarkably stable and decodable. Given that the model’s core knowledge‑boundary neural pathways remain relatively intact, conventional approaches~\cite{zhang2023r, cheng2024can,xu2024sayself} that require extensive honesty-recovery datasets for global parameter adjustment may be unnecessarily resource-intensive. In the following section, we introduce a more targeted framework that identifies honesty-critical neurons, which are pivotal for articulating the model's self-knowledge of its knowledge boundaries, and adjust these neurons to efficiently restore the truthfulness of fine-tuned LLMs.

\section{Honesty-Critical Neurons Restoration}

Building on the above findings, we design a targeted remedy rather than a global parameter overhaul. The Honesty‑Critical Neurons Restoration (HCNR) framework (Figure \ref{fig:overview}) focuses on the neurons that (i) govern the expression of honesty, (ii) exert little influence on downstream tasks, and (iii) are most disrupted by SFT. By first reverting these neurons to their pre‑training states and then applying a Hessian‑guided compensation, HCNR restores truthful responses with negligible impact on task performance.

\subsection{Honesty-Critical Neurons Recognition}
\noindent \textbf{Intra-layer Sensitivity Assessment.} To systematically identify neurons critical for honesty expression, we begin by assessing neurons' importance through quantifying how parameter perturbations affect honesty task loss. Inspired by the OBD pruning method~\cite{lecun1989optimal}, our approach evaluates the impact of SFT-induced parameter perturbation $\delta\theta$ on honesty task loss $\mathcal{L}_{\mathrm{hon}}$ for an LLM parameterized by $\theta$. This relationship can be expressed via the second-order Taylor expansion:
\begin{equation}
    \Delta\mathcal{L}_{\mathrm{hon}}=\nabla_\theta\mathcal{L}_{\mathrm{hon}}\cdot\delta\theta+\frac{1}{2}\delta\theta^\top H\delta\theta+O(\|\delta\theta\|^3),
\end{equation}
where $H=\nabla^{2}_{\theta}\mathcal{L}_{\mathrm{hon}}$ represents the Hessian matrix of the loss function. Given that the pre-trained model has converged to a local optimum, the gradient term $\nabla_\theta\mathcal{L}_{\mathrm{hon}}\approx0$, thereby reducing the loss change to:
\begin{equation}
    \Delta\mathcal{L}_{\mathrm{hon}}\approx\frac{1}{2}\delta\theta^\top H\delta\theta.
\end{equation}

\begin{theorem}
Under the assumptions that: (1) At each SFT step, the parameter increment $\delta\theta$ has zero mean and an isotropic covariance: $\mathbb{E}[\delta\theta] = 0,\ \mathbb{E}[\delta\theta\delta\theta^T]=\sigma^2I_d$, (2) and given sufficient observational data, we have:
\begin{equation}
    \mathbb{E}[\Delta\mathcal{L}_{\mathrm{hon}}]\approx\frac{1}{2}\sum_{i} \sigma^{2}F_{ii}\ \ \propto\ \ F_{ii},
\end{equation}
where $F_{ii}$ denotes the diagonal element of the Fisher Information Matrix (FIM). We approximate $F_{ii}$ with the empirical mean of gradients of the log-likelihood.
\end{theorem}

According to Proposition 1 (proved in Appendix E), diagonal elements of the FIM serve as unbiased estimators of neuron importance. Specifically, for the $k$-th neuron in layer $j$ weights $W_j\in\mathbb{R}^{d'\times d}$ (where $d'$ and $d$ are the hidden dimensions), we define its importance $s_{j,k}$ on task $D$ as:
\begin{equation}
\label{eq:4}
    s_{j,k}=\mathbb{E}_{(x,y)\sim D}[(\partial_{W_{j,k}}\mathcal{L})^2].
\end{equation}

To measure the importance of neuron $W_{j,k}$ for both honesty and downstream tasks, we calculate its respective importance scores $s^{\text{hon}}_{j,k}$ and $s^{\text{task}}_{j,k}$ using minimal honesty data $D^{\text{hon}}$ and downstream task data $D^{\text{task}}$. To prioritize neurons that are crucial for honesty yet secondary for downstream tasks, we define the priority $r_{j,k}$ of $W_{j,k}$:
\begin{equation}
\label{eq:5}
    r_{j,k}=s^{\text{hon}}_{j,k}\cdot log\frac{s^{\text{hon}}_{j,k}}{s^{\text{task}}_{j,k}}.
\end{equation}

Higher $r_{j,k}$ values indicate neurons with high honesty contribution and low downstream task contribution. We aim to protect such neurons given their critical role in honesty and their non-interference with downstream task performance. Specifically, we rank all neurons in $W_j$ by priority $r_{j,k}$ in descending order and select the top $d'\times R_{IW}$ neurons to form the candidate neuron set $A_{j}^{\text{neuron}}$, where $R_{IW}$ represents the in-weight ratio hyperparameter.

\noindent \textbf{Cross-layer Perturbation Analysis.} While intra-layer analysis identifies honesty-related neurons, SFT induces uneven perturbation intensity across layers due to LLMs' hierarchical specialization, which means indiscriminate layer protection would overly constrain downstream performance. Therefore, we prioritize layers with substantial SFT perturbation through relative weight displacement $d_j$:
\begin{equation}
\label{eq:6}
    d_j=\frac{\parallel(W_j-W_j^{\prime})\odot M_j\parallel_2}{\parallel W_j\odot M_j\parallel_2},\ \ \ j=1,...,L,
\end{equation}
where $W_j$ and $W_j^{\prime}$ represent the $j$-th layer weights before and after fine-tuning, and $M_j$ denotes a binary mask constructed from $A_{j}^{\text{neuron}}$, $M_{j,k:}=1$ if the $k$-th neuron is a candidate (i.e., $k\in A^{neuron}_j$), and 0 otherwise. This metric reflects the perturbation intensity of SFT on candidate neurons: higher $d_j$ values indicate that honesty-related neurons in layer $j$ have been more drastically altered by SFT, requiring priority protection. We rank layers by $d_j$ in descending order and select the top $L\times R_{CW}$ layers to form the candidate set $A^\mathrm{layer}$, where the hyperparameter $R_{CW}\in(0,1)$ is the cross-weight ratio and it controls the proportion of layers requiring focused protection.

\noindent \textbf{Neuron Restoration.} Following the identification of critical layers and neurons, we combine candidate layers $A^{\text{layer}}$ and per-layer candidate neuron sets $A_{j}^{\text{neuron}}$ to identify the final honesty-critical neuron set $A^{\text{hc}}$:
\begin{equation}
    A^{\text{hc}}=\{(j,k)|j\in A^{\text{layer}}\ and\ k\in A_{j}^{\text{neuron}}\},
\end{equation}
these neurons are crucial for honesty expression and significantly perturbed during SFT, causing the model's knowledge boundary awareness to be ``masked" at the output stage. To restore fine-tuned model honesty, we revert neurons in $A^{\text{hc}}$ to their pre-training states while keeping remaining neurons $A^{\text{task}}$ (downstream task neurons) unchanged.

\subsection{Honesty Compensation}
However, simply reverting honesty-critical neurons creates a new challenge: since SFT involves coordinated parameter updates across all layers, the neural activation pathways continuously evolve. Restoring only honesty-critical neurons causes misalignment with downstream task neurons $A^{\text{task}}$, leading to a rebound of the loss of honesty tasks. To address this issue, we introduce Honesty Compensation.

Specifically, we denote $\boldsymbol{\mathrm{W}}^{\mathrm{orig}}$, $\boldsymbol{\mathrm{W}}^{\mathrm{sft}}$, and $\boldsymbol{\mathrm{W}}^{\mathrm{hc}}$ as the pre-training, fine-tuned, and post-restoration parameters, respectively. Let $X_{hon}$ denote the inputs of examples from $D^{hon}$. Our objective is to minimize the activation difference between $\mathrm{W}^{\mathrm{hc}}$ and $\mathrm{W}^{\mathrm{orig}}$ on honesty tasks:
\begin{equation}
    d_{hon} = ||\boldsymbol{\mathrm{W}}^{\mathrm{hc}}X_{hon}-\boldsymbol{\mathrm{W}}^{\mathrm{orig}}X_{hon} ||_2^2.
\end{equation}

Inspired by OBS~\cite{hassibi1993optimal}, we present Proposition 2 (proved in Appendix F), which derives the optimal compensation vector for honesty-critical neurons that exactly counteracts the increase in $d_{hon}$ caused by their misalignment with downstream-task neurons.

\begin{theorem}
Consider a layer parameter $W_j^{\text{orig}}$ in the pre-training model. The increment introduced by SFT at position $k$ is $\delta w_{j,k}=W^{\text{sft}}_{j,k}-W^{\text{orig}}_{j,k}$, which increases $d_{hon}$. To compensate, we apply the adjustment $c_{j,k}$ to the restored honesty-critical parameters:
\begin{equation}
\label{eq:9}
    c_{j,k} = \frac{W^{\text{sft}}_{j,k}-W^{\text{orig}}_{j,k}}{[H^{-1}]_{kk}} \cdot H_{:,k}^{-1},
\end{equation}
where $H=\nabla_{W^{\text{orig}}_j}^2 d_{hon}$ is the Hessian matrix of $d_{hon}$, computed on $D^{\text{hon}}$. 
\end{theorem}

Thus, this principle culminates in our final weight update rule, which operates conditionally. To preserve task capabilities, downstream task neurons ($A^{\text{task}}_j$) retain their fine-tuned values. Concurrently, each honesty-critical neuron ($i \in A^{\text{hc}}_j$) is reverted to its pre-training state and then adjusted by an aggregated compensation term.

\subsection{Overview of HCNR}

Our Honesty-Critical Neurons Restoration (HCNR) framework operates in two sequential stages to restore honesty while preserving downstream task performance.  

\noindent \textbf{Stage 1: Recognition of Honesty-Critical Neurons.} We first compute neuron importance scores and priorities for both honesty and downstream tasks using Equations (\ref{eq:4}) and (\ref{eq:5}) on minimal labeled datasets. Neurons are ranked by their priorities in descending order, with the top $R_{IW}$ proportion selected as candidate neurons within each layer. Subsequently, we quantify the perturbation intensity of these candidates during SFT using Equation (\ref{eq:6}) and identify the $R_{CW}$ proportion of layers with the most severe perturbations as honesty-critical neurons. Finally, we restore these identified neurons to their pre-training states while preserving the fine-tuned parameters of downstream task neurons.

\noindent \textbf{Stage 2: Honesty Compensation.} To address the misalignment between restored honesty-critical neurons and downstream task neurons, we compute compensation vectors using Equation (\ref{eq:9}) and integrate them into the restored parameters. The final HCNR weights are determined by:

\begin{equation}
W_{j,i}^{\text{HCNR}} =
\begin{cases}
W_{j,i}^{\text{orig}} + [\sum_{k \in A^{\text{task}}_j} c_{j,k}]_i &\text{if } i\in A^{\text{hc}}_j \\
W_{j,i}^{\text{sft}} & \text{if } i \in A^{\text{task}}_j,
\end{cases}
\label{eq:final_weights}
\end{equation}
where $W_{j,i}^{\text{HCNR}}$ denotes the final weight of the $i$-th neuron in layer $j$, and $[\cdot]_i$ extracts the $i$-th component of the vector.

\begin{table*}[htbp]
\setlength{\tabcolsep}{3.5pt} 
\centering
\renewcommand{\arraystretch}{1.3}
\small 
\begin{tabular}{c|c|cc|cc|cc|cc|cc|c}
\toprule
\multirow{2}{*}{\textbf{Datasets}}
& \multirow{2}{*}{\textbf{Methods}}
& \multicolumn{2}{c|}{\textbf{FalseQA}}
& \multicolumn{2}{c|}{\textbf{NEC}}
& \multicolumn{2}{c|}{\textbf{RefuNQ}}
& \multicolumn{2}{c|}{\textbf{KUQ}}
& \multicolumn{2}{c|}{\textbf{SelfAware}}
& \multicolumn{1}{c}{\textbf{Domain}}
\\
& & F1$\uparrow$ & RF $\Delta \uparrow$
& F1$\uparrow$ & RF $\Delta \uparrow$
& F1$\uparrow$ & RF $\Delta \uparrow$
& F1$\uparrow$ & RF $\Delta \uparrow$
& F1$\uparrow$ & RF $\Delta \uparrow$
& Accuracy $\uparrow$
\\
\midrule

\multirow{7}{*}{\textbf{HotpotQA}}
&Fine-tuned& 56.51 & +23.34  & 35.46 & +13.78  &  32.43 & +17.10   & 68.50 & +50.06 & 67.01 &	+51.14  & 30.65\\
\cline{2-13}
&ICL& 16.23 &	+7.06   & 8.64 &+3.31& 	19.84 &	+9.33  & 42.89 	&+27.09  & 40.15 & +24.84  & 30.65\\
&RAIT & 68.59& +20.93  & 68.28 &	+8.94   & 71.21 &+32.49  & 80.38 &+53.82  & 64.46 &	+50.83  & 27.05\\
&Rehearsal & 67.05 	&+2.28  & 67.01 &+1.35   & 66.86 	&+6.84  & 69.31 &	+11.75  & 48.51 &+6.25  & 29.15\\
&DPO & \textbf{69.12} & \textbf{+25.75}  & 69.52 &+14.44  & \textbf{72.91} &+40.17  & 80.96 &+55.13  & 64.76 &	+51.59  &29.00\\
&ORPO & 65.83 & +23.68  & 70.03 &+23.97  & 71.26 &\textbf{+45.16}  & 79.21 & +55.92  & 65.21 &+50.97  &29.60\\
&\textbf{HCNR(ours)} & 68.30 & +17.60  & \textbf{71.90} 	&\textbf{+35.80}  & 71.70 &+38.00  & \textbf{82.90} &\textbf{+62.00}  & \textbf{69.40} &\textbf{+59.80}  &30.30\\

\midrule
\multirow{7}{*}{\textbf{MedMCQA}}
&Fine-tuned &  58.48& \textbf{+28.63}  & 45.09 & +20.61  & 52.43  &  +29.58  & 67.49 & +48.08 & 62.70 & +45.87  & 10.30\\

\cline{2-13}
&ICL& 32.27  &+14.46& 42.60 & +23.61& 58.95	 &	+36.44  & 58.15 	& +39.11 & 53.32 & +35.82  &10.30 \\
&RAIT & \textbf{69.59} & +22.75  & 69.37 &	+12.99   &69.26  &+25.63  &71.99  &+28.38  & 54.73 &	+28.71  & 9.60\\
&Rehearsal &  67.33	& +3.68 & 67.32 & +2.70  &  67.66	& +12.05 & 69.54 &	+12.88 & 49.31 	& +9.23 & 10.35 \\
&DPO & 67.98 & +24.95 & 70.57 & +19.12 & \textbf{71.61} & +38.53 & 72.76 &+34.46 & 54.85 &	+30.87  & 9.49 \\
&ORPO & 65.22 & +28.20 & 69.19 & +34.54 & 66.58 & +40.47 & 80.17 & +61.06 & 69.22 & +55.90& 10.09\\
&\textbf{HCNR(ours)} & 69.40 & +25.70 	& \textbf{71.00} & \textbf{+39.70}  & 70.70 & \textbf{+40.60} & \textbf{83.00} & \textbf{+65.20} & \textbf{71.00} & \textbf{+59.90} & 10.33\\

\bottomrule
\end{tabular}
\caption{Comparison of F1 score (\%), Refusal $\Delta$ (RF $\Delta$ \%), and domain accuracy (\%) between our method and baseline methods on multiple honesty benchmarks and domain tasks. Results are evaluated on the recovery of the Llama-3.1-8B-Instruct model fine-tuned on the HotpotQA and MedMCQA datasets.}
\label{tab:effectiveness}
\end{table*}

\begin{figure}[t]
    \centering
    \includegraphics[width=\linewidth]{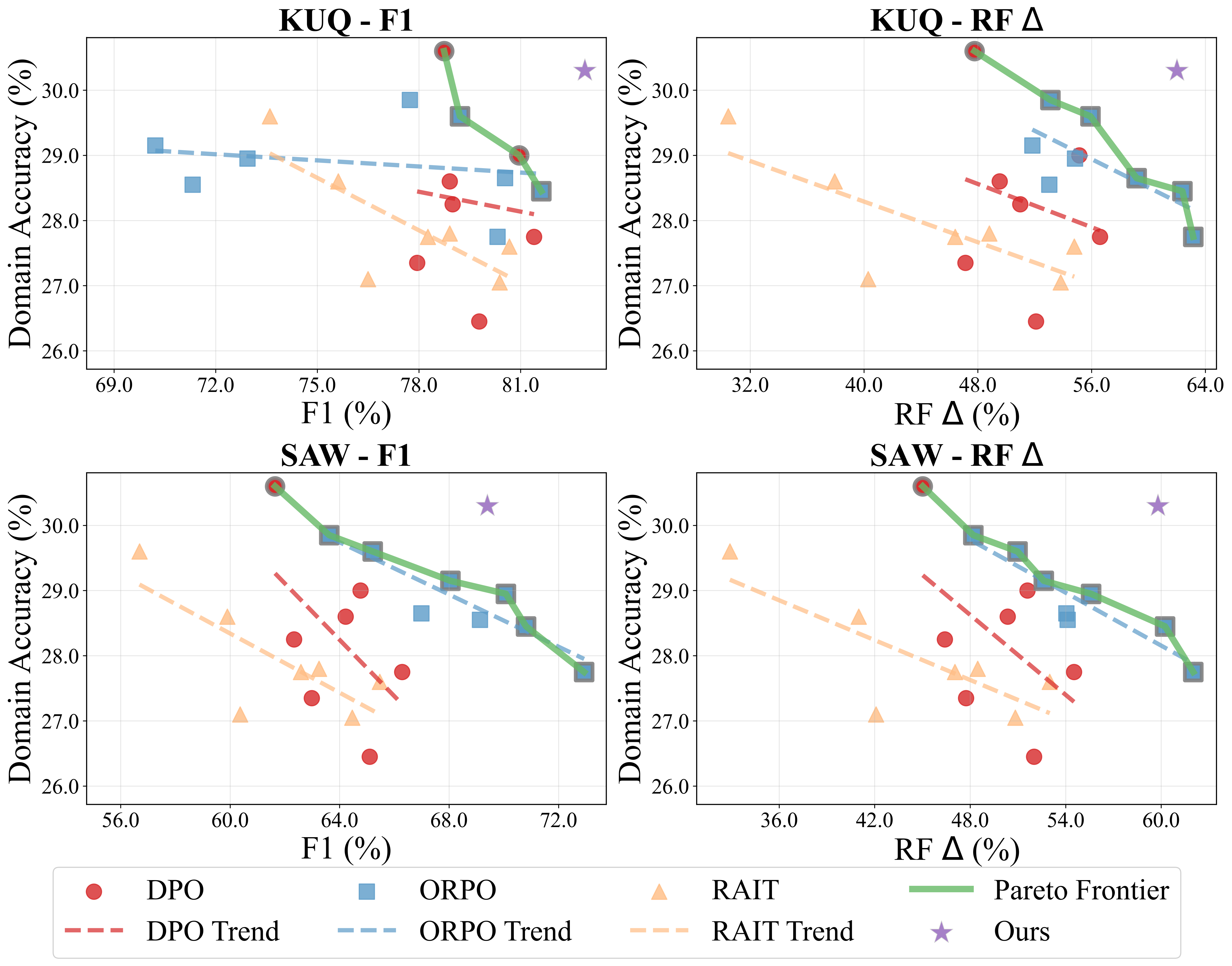}
    \caption{Task-honesty trade-off comparison between our method and baselines on SelfAware and KUQ datasets. HCNR outperforms all baselines' Pareto frontier, achieving a superior task-honesty balance.}
    \label{fig:balance}
\end{figure}

\section{Experiments}

In this section, we evaluate HCNR along four key dimensions: (1) effectiveness of honesty restoration, (2) the balance between domain-specific performance and honesty, (3) deployment efficiency, and (4) generalizability across different LLM families and training paradigms.

\subsection{Experimental Setup}
\noindent \textbf{Models}. To ensure broad applicability across different model families and architectures, we evaluate our HCNR framework on multiple open-source models, including Llama-3.1-8B-Instruct, Llama-3-8B-Instruct, Qwen3-8B-Instruct, Qwen2-7B-Instruct, and Mistral-7B-Instruct. 

\noindent \textbf{Fine-tuning Datasets}. For fine-tuning, we utilize four question-answering datasets to simulate domain-specific fine-tuning: factual QA datasets including HotpotQA~\cite{yang2018HotpotQA}, and Natural Questions~\cite{kwiatkowski2019natural}, and specialized domain datasets including MedMCQA~\cite{pal2022MedMCQA}, and BioASQ~\citep{krithara2023bioasq} (details are shown in Appendix A). Models undergo Low-Rank Adaptation (LoRA)~\cite{hu2022lora} or full fine-tuning with hyperparameters detailed in Appendix G. We evaluate downstream task performance using accuracy on task-specific test sets or mean ROUGE-L scores for specific tasks (BioASQ).

\noindent \textbf{Honesty Evaluation}. Our honesty evaluation suite includes SelfAware (SAW)~\cite{yin2023large}, Known-Unknown Questions (KUQ)~\cite{amayuelas2023knowledge}, and three datasets from UnknownBench~\cite{liu2023examining}: FalseQA, Non-existent Concepts (NEC), and Refusal-inducing Natural Questions (RefuNQ). Dataset details are provided in Appendix A. Following established conventions in honesty evaluation~\cite{li2024survey,yang2024alignment}, we employ the following metrics: (1) \textit{F1 Score}: Balances precision and recall for identifying unanswerable questions (positive class). (2) \textit{Refusal $\Delta$ (RF $\Delta$)}: Difference between refusal rates on unanswerable versus answerable questions.

\noindent \textbf{Baseline Methods}. We compare HCNR to five representative honesty-recovery strategies. Following prior work in honesty alignment~\cite{cheng2024can,zhang2023r}, methods (2)–(5) use an IDK (I don't know) dataset (details are shown in Appendix B). (1) \textit{In-Context Learning (ICL)}. Few-shot prompts include exemplars that model appropriate uncertainty. (2) \textit{RAIT}. Standard SFT of the fine-tuned model on the IDK dataset to directly teach IDK responses. (3) \textit{Rehearsal}. During downstream SFT, mix a small portion of IDK data with downstream task data to preserve IDK behavior while learning the domain task. (4) \textit{DPO.} Warm up the degraded model with one epoch of SFT on IDK, then apply Direct Preference Optimization~\cite{rafailov2023direct} using preference-formatted IDK pairs. (5) \textit{ORPO.} Apply Odds Ratio Preference Optimization~\cite{hong2024orpo} directly on IDK data; its joint supervised-plus-preference loss removes the need for a warm-up phase.

\noindent \textbf{Implementation Details}. In HCNR, both the honesty and task datasets are set to 128 samples ($|D^{\text{hon}}|=|D^{\text{task}}|=128$), with the in-weight ratio $R_{IW}=0.5$ and cross-weight ratio $R_{CW}=0.4$. We follow existing works~\cite{cheng2024can,zhang2023r} to construct the honesty dataset $D^{\text{hon}}$ (details are provided in Appendix B). All experiments repeat
three times, and the average results are recorded.

\begin{table}[]
\centering
\small
\setlength{\tabcolsep}{2.5pt}
\begin{tabular}{ccccrcc}
\toprule 
\textbf{Dataset}&\textbf{Methods} & \textbf{Size$\downarrow$} & \textbf{Ratio $\downarrow$}& \textbf{Time$\downarrow$}& \textbf{F1}$\uparrow$ & \textbf{RF} $\Delta \uparrow$\\
\midrule
\multirow{4}{*}{\textbf{Hot}}
&RAIT & 5000 &100\%& 8.76 min & 70.58 & +33.40\\
&DPO & 5000 &100\%& 42.78 min& 71.45 &+37.41\\
&ORPO & 9000 &100\%&  30.97 min&70.31&+39.94\\
&\textbf{HCNR} & \textbf{256} & \textbf{20\%} &\textbf{3.93 min}&\textbf{72.84} &\textbf{+42.64}\\
\midrule
\multirow{4}{*}{\textbf{Med}}
&RAIT & 5000&100\%& 8.81 min &66.99  & +23.69\\
&DPO &  3000 & 100\%&  24.28 min& 67.55  &+29.59\\
&ORPO &  5000 &100\%& 17.21 min & 70.08  &+44.03\\
&\textbf{HCNR} & \textbf{256} & \textbf{20\%} &\textbf{3.67 min}&\textbf{73.02} &\textbf{+46.22}\\
\bottomrule
\end{tabular}
\caption{Comparison of additional data volume (Size), parameter modification ratio (Ratio), and time overhead (Time) between HCNR and baselines. Results are evaluated on the recovery of Llama-3.1-8B-Instruct fine-tuned on the HotpotQA (Hot) and MedMCQA (Med). F1 and RF \text{$\Delta$} are the average values across five honesty benchmarks. HCNR achieves optimal results with maximal efficiency.}
\label{tab:time}
\end{table}

\subsection{Main Results}

\noindent \textbf{Effectiveness of HCNR.} We evaluate our approach for restoring honesty in models compromised by HotpotQA and MedMCQA fine-tuning, comparing it against five baseline methods. As shown in Table \ref{tab:effectiveness}, HCNR achieves superior performance across most honesty benchmarks, effectively restoring the honesty of fine-tuned models while maintaining their downstream task capabilities. Although RL-based methods (DPO and ORPO) demonstrate competitive results on a few benchmarks, they substantially compromise downstream task performance. Notably, the ICL prompting approach exhibits poor honesty performance, which we attribute to instruction-tuning's detrimental effects on LLMs' in-context learning capabilities~\cite{wang2024loss}. This finding underscores the effectiveness of our HCNR framework for honesty restoration.

\noindent \textbf{Balance between domain task and honesty}. In practice, effective honesty recovery methods need to balance downstream task performance and honesty, achieving an optimal trade-off between them. For baseline post-alignment methods, the IDK dataset size represents a critical parameter that impacts this balance. To systematically evaluate this trade-off, we compare HCNR against baseline methods across varying dataset sizes. Note that we exclude the Rehearsal method from this analysis due to its poor RF $\Delta$ performance.

We conduct this analysis using Llama-3.1-8B-Instruct, which is fine-tuned on HotpotQA. We assess honesty using the KUQ and SelfAware benchmarks. As illustrated in Figure \ref{fig:balance}, our method successfully achieves the desired equilibrium between task performance and honesty, while existing baselines consistently fail to maintain this balance. Specifically, baseline methods (DPO, ORPO, and RAIT) exhibit a persistent negative correlation between task performance and honesty. Our approach, however, consistently outperforms the Pareto frontier established by these baselines, demonstrating superior capability in addressing the fundamental challenge of preserving domain-specific task performance while enhancing model honesty.

\begin{table}[t]
\centering
\small
\setlength{\tabcolsep}{3.5pt}
\begin{tabular}{cccccc}
\toprule 
\textbf{Dataset} & \textbf{Stage 1} & \textbf{Stage 2}& \textbf{F1}$\uparrow$ & \textbf{RF} $\Delta \uparrow$&\textbf{Domain}$\uparrow$ \\
\midrule
\multirow{5}{*}{\textbf{HotpotQA}}
& Random& Ours &65.44 & +36.31 &  29.60 \\
& w/o Task& Ours & 70.43 & +33.24&  28.30\\
& Ours& w/o Com & 65.96 & +33.09 &  30.37 \\
& Random& w/o Com& 54.21& +23.04& 29.70\\
&\textbf{Ours} &\textbf{Ours} & \textbf{72.84} & \textbf{+42.64}&  \textbf{30.30}\\
\midrule
\multirow{5}{*}{\textbf{MedMCQA}}
& Random& Ours & 67.39 & +40.40 &  10.30 \\
& w/o Task& Ours & 72.74 & +43.53&  9.32\\
& Ours& w/o Com& 65.68& +41.77& \textbf{10.40}\\
& Random& w/o Com& 58.70& +35.07& \textbf{10.40}\\
&\textbf{Ours} & \textbf{Ours} &\textbf{73.02} & \textbf{+46.22}&  10.33\\
\bottomrule
\end{tabular}
\caption{Ablation study evaluating key components of our two-stage method. We test stage 1 modifications (Random, w/o Task) and stage 2 changes (w/o Com) against our original method (``Ours"). Performance measured by average F1, RF \text{$\Delta$} across five honesty benchmarks, and domain accuracy.}
\label{tab:identi_obs}
\end{table}

\noindent \textbf{Efficiency of HCNR.} We evaluate HCNR's computational efficiency against established baselines using the Llama-3.1-8B-Instruct fine-tuned on HotpotQA and MedMCQA, with experiments conducted on an Nvidia A800-80GB GPU. For a fair comparison, baseline methods are tuned to their optimal configurations by varying their training data size to achieve the best trade-off between task performance and honesty. The results, presented in Table \ref{tab:time}, highlight HCNR's superior efficiency. It achieves optimal performance with a mere 256 data samples (128 each for $D^{\text{hon}}$ and $D^{\text{task}}$) while modifying just 20\% of the model's parameters. Furthermore, as a training-free method, HCNR is inherently more efficient, delivering at least a 2.23x speedup over all baselines. These results confirm that HCNR offers a practical path to honesty recovery that is efficient in both data and time, making it viable for real-world deployment.

\noindent \textbf{More experiments based on other LLMs and FFT.} We also evaluate it on Llama-3-8B-Instruct, as well as Qwen and Mistral series models. Additionally, we conduct experiments using full fine-tuning (FFT) to further assess HCNR across training paradigms. The results are shown in Appendix D, demonstrating that HCNR consistently achieves honesty improvements and high efficiency across all these settings. 

\subsection{Ablation Study}
\noindent \textbf{Ablation Studies on HCNR's Components.} To validate the effectiveness of our proposed HCNR framework, we conduct ablation studies on both core components. For Stage 1 (honesty-critical neurons recognition), we compare our method against Random selection (randomly select neurons) and w/o Task variants (using $r_{j,k} = s_{j,k}^{hon}$ without task considerations).  The number of neurons selected is the same across all experimental settings. Table \ref{tab:identi_obs} shows HCNR achieves F1 improvements over Random selection, while the w/o Task variant causes domain task performance drops, demonstrating the necessity of critical neuron identification. For Stage 2 (honesty compensation), comparison with the w/o Compensation (w/o Com) variant reveals the critical importance of honesty compensation. When both stage components are simultaneously ablated (Random and w/o Com), honesty recovery becomes substantially impaired. These results validate both the precision of our neuron identification and the necessity of a compensation mechanism for effective honesty enhancement while preserving task-specific capabilities.
\begin{figure}
    \centering
    \includegraphics[width=\linewidth]{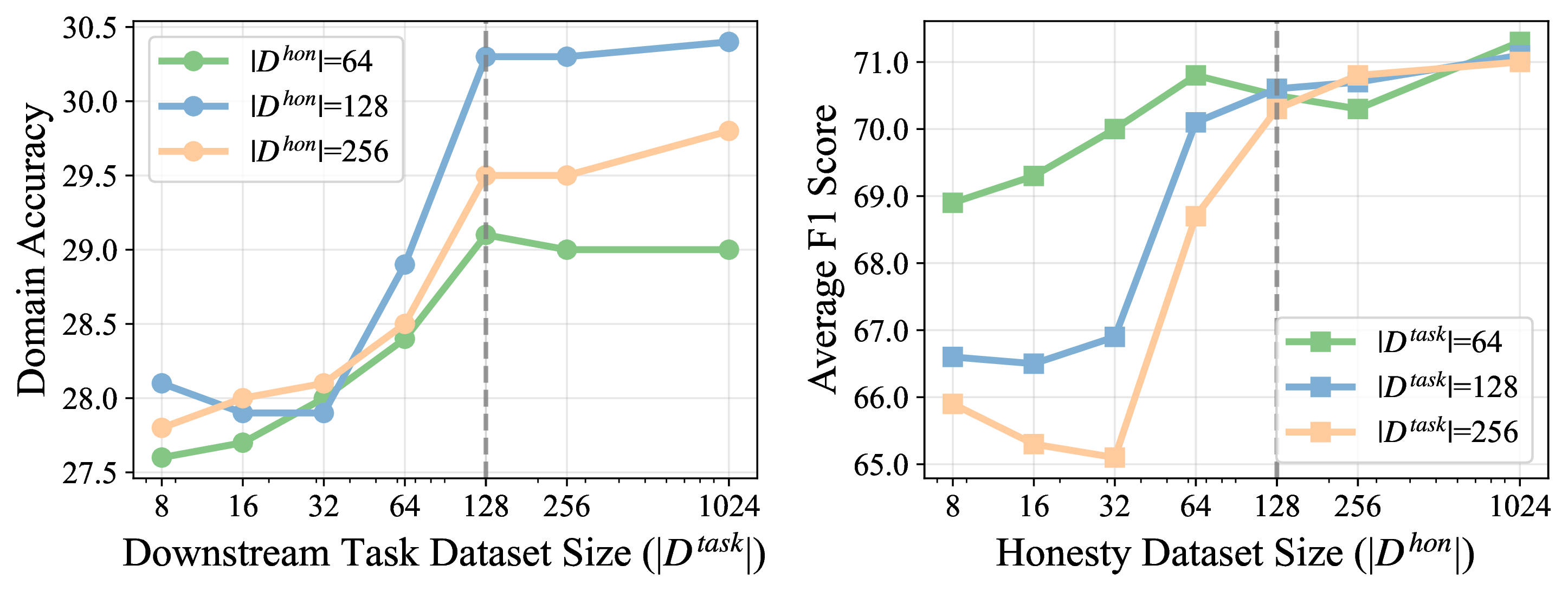}
    \caption{Ablation study on the required dataset sizes \textbf{$|D^{task}|$} and \textbf{$|D^{hon}|$} for HCNR. We vary the dataset sizes and record the recovered models' average F1 scores on UnknownBench and in-domain task accuracy.}
    \label{fig:hyper_dsize}
\end{figure}

\noindent \textbf{Influence of $|D^{task}|$ and $|D^{hon}|$.} To evaluate the data efficiency of the HCNR framework, we conduct ablation studies on the sizes of the two required datasets: $D^{hon}$ and $D^{task}$. We systematically vary these dataset sizes while recovering HotpotQA fine-tuned models under the same settings. As Figure \ref{fig:hyper_dsize} shows, both honesty restoration and downstream accuracy plateau with only $\approx128$ examples. These findings confirm that the HCNR framework achieves optimal recovery performance for both honesty restoration and task preservation using remarkably few examples, highlighting its practical efficiency.

\begin{figure}
    \centering
    \includegraphics[width=\linewidth]{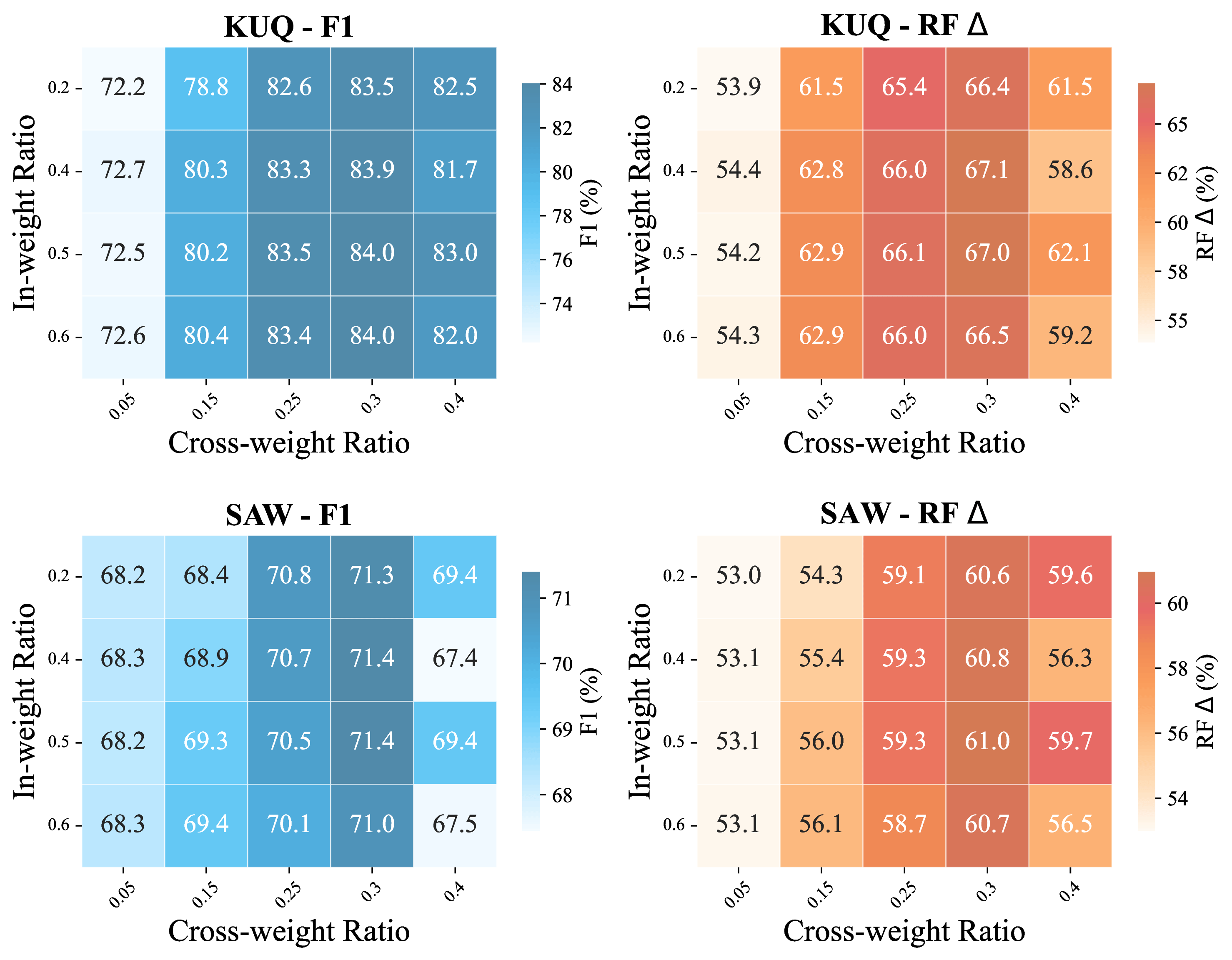}
    \caption{Ablation study of in-weight ratio $R_{IW}$ and cross-weight ratio $R_{CW}$ hyperparameters on honesty recovery for the fine-tuned model on the HotpotQA dataset.}
    \label{fig:hyper_abala}
\end{figure}

\noindent \textbf{Influence of $R_{IW}$ and $R_{CW}$.} To investigate the impact of $R_{IW}$ and $R_{CW}$ hyperparameters on honesty recovery, we conduct ablation studies on honesty-impaired models, evaluating performance on KUQ and SelfAware datasets. As shown in Figure \ref{fig:hyper_abala}, both F1 score and Refusal $\Delta$ improve progressively with increasing hyperparameter values. However, $R_{IW}$ exhibits rapid saturation with diminishing marginal gains, while $R_{CW}$ achieves optimal performance around 0.3, with further increases causing performance degradation.

\section{Related Works}
\subsection{LLMs' Honesty and Self-Knowledge}
Honesty-enhancing methods can be categorized into two approaches: refusal-aware instruction tuning (RAIT) and reinforcement learning~\cite{li2024survey}. These approaches all require constructing large-scale, meticulously designed training datasets followed by retraining. RAIT methods~\cite{zhang2023r} evaluate LLMs' knowledge boundaries by sampling and modifying labels for subsequent SFT. Reinforcement learning approaches~\cite{cheng2024can} enhance honesty through constructed preference data and reward modeling. Both approaches inevitably cause catastrophic forgetting and performance degradation on in-domain tasks. Moreover, all existing approaches overlook the underlying mechanisms of honesty deterioration, focusing solely on post-hoc correction.

\subsection{Knowledge and Capability Neurons}
The concept of knowledge and capability neurons has emerged as a promising approach for attributing and localizing model behavior~\cite{niu2024does,ferrando2024primer}. Existing research has demonstrated that specific neurons correlate with various knowledge concepts~\cite{dai2021knowledge,shi2024ircan}. Recent advances have extended this line of inquiry to critical capabilities, revealing associations between specific neurons and safety~\cite{yi2025nlsr}, and confidence regulation~\cite{stolfo2024confidence}. Building on the demonstrated success of neuron-level analysis across various domains, this work investigates honesty-related neurons to address the gap in understanding how large language models process honesty at the neuronal level.

\section{Conclusion}
This work addresses honesty degradation in fine-tuned LLMs that leads to confident fabrications. Mechanistically, we reveal that dishonesty is spurious: while SFT alters expression behavior, inner knowledge boundary representations remain stable. This insight enables our Honesty-Critical Neurons Restoration (HCNR), which restores truthful behavior by selectively reverting honesty-critical neurons to pre-trained states and employing Hessian-guided compensation. Extensive experiments demonstrate that HCNR achieves substantial honesty recovery with minimal overhead while preserving task performance, providing an efficient solution for trustworthy AI in high-stakes domains.

\section{Acknowledgements}
The work is supported by the grants from the Natural Science Foundation of China (62225202, 62202029), and Young Elite Scientists Sponsorship Program by CAST (No. 2023QNRC001). We owe sincere thanks to all authors for their valuable efforts and contributions. The corresponding author is Tianyu Chen and Haoyi Zhou.

\bibliography{main}
\end{document}